\documentclass[11pt]{article}
\usepackage[preprint]{acl}
\usepackage{times}
\usepackage{latexsym}
\usepackage[T1]{fontenc}
\usepackage[utf8]{inputenc}
\usepackage[nopatch=footnote]{microtype}
\usepackage{inconsolata}
\usepackage{graphicx}
\usepackage{pgf}
\usepackage{amsmath} 
\usepackage{tikz}
\usepackage{adjustbox}
\usepackage{framed}
 \usepackage{threeparttablex}
 \usepackage{enumitem}
\usepackage{tikz}
\usepackage{svg}
\usepackage{tikz-cd}
\usepackage{pgfplots}
\usepackage{amsfonts}
\usepackage{pdflscape}
\pgfplotsset{compat=1.18}
\newcommand{\mathdefault}[1][]{}
\usepackage{setspace}
\usepackage{eurosym}
\usepackage{silence}
\usepackage{pdfpages}
\usepackage{placeins}

\title{An Evaluation of Explanation Methods for Black-Box Detectors of Machine-Generated Text}
\author{
 \textbf{Loris Schoenegger\textsuperscript{1,2}},
 \textbf{Yuxi Xia\textsuperscript{1,2}},
 \textbf{Benjamin Roth\textsuperscript{1,3}}
\\
\\
 \textsuperscript{1}Faculty of Computer Science, University of Vienna, Vienna, Austria\\
 \textsuperscript{2}UniVie Doctoral School Computer Science, University of Vienna, Vienna, Austria\\
  \textsuperscript{3}Faculty of Philological and Cultural Studies, University of Vienna, Vienna, Austria\\
\\
 \small{
   \textbf{Correspondence:} \href{mailto:loris.schoenegger@univie.ac.at}{loris.schoenegger@univie.ac.at}
 }
}
\begin{document}
\maketitle
\begin{abstract}
The increasing difficulty to distinguish language-model-generated from human-written text has led to the development of detectors of machine-generated text (MGT).
However, in many contexts, a black-box prediction 
is not sufficient, it is equally important to know \emph{on what grounds} a detector made that prediction. 
Explanation methods that estimate feature importance 
promise to provide indications of which parts of an input are used by classifiers for prediction.
However, these are typically evaluated with simple classifiers and tasks that are intuitive to humans.
To assess their suitability beyond these contexts, this study conducts the first systematic evaluation of explanation quality for detectors of MGT.
The dimensions of \textit{faithfulness} and \textit{stability} are evaluated with five automated experiments, and \textit{usefulness} is assessed in a user study. We use a dataset of ChatGPT-generated and human-written documents, and pair predictions of three existing language-model-based detectors with the corresponding SHAP, LIME, and Anchor explanations.
We find that SHAP performs best in terms of faithfulness, stability, and in helping users to predict the detector's behavior. 
In contrast, LIME, \emph{perceived} as most useful by users, scores the worst in terms of user performance at predicting detector behavior.
\end{abstract}
\section{Introduction}
Large language models, such as ChatGPT, produce output that is often virtually indistinguishable from human-written text. Their ability to generate human-like text at an unprecedented scale allows for new forms of phishing, disinformation campaigns, and academic fraud~\cite{crothers_threat_2023}.
Recent work has proposed language-model-based detection methods for machine-generated text \cite[MGT;][]{solaiman_release_2019,guo_how_2023,mitchell_detectgpt_2023}.
These operate as black-box detectors: they provide no explanations for their decisions. 
This is insufficient for applications that demand additional evidence, or when wrong decisions affect people, 
as would be the case for detecting MGT in academia.

To address this, \textbf{explanation methods} have been applied to such detectors \cite{mosca_distinguishing_2023, liu_check_2023, yu_cheat_2023}, with the majority of papers using SHAP \cite{lundberg_unified_2017} or LIME \cite{ribeiro_why_2016}, two methods that assign feature importance scores to individual tokens in the input (higher importance scores suggest stronger impact on the detector's decision). We find that Anchor \cite{ribeiro_anchors_2018}, a method that generates rule-based explanations (if-then rules, e.g.,:\textit{ If the document contains the word "delve", the detector will predict "machine" in 70\% of cases}), can also be applied in this setting.
All three methods produce local explanations that explain single predictions (MGT or not) by \textbf{locating relevant elements in the input} (words in the document that influenced the prediction).

Previously, LIME and SHAP have been used together with detectors of MGT without verifying the quality of the resulting explanations.
This is problematic as prior evaluations relied on tasks such as sentiment analysis, which have comparatively simple feature sets that are more intuitive to users.
In this work, we rely on MGT detection as a case study for tasks characterized by \textbf{understudied feature sets}. The novelty of our work is an evaluation that goes \textbf{beyond typical evaluation tasks}, assessing explanations in terms of three important aspects of quality, defined as follows:
\begin{enumerate}[nosep, topsep=0pt, label=(\arabic*), left=0pt]
    \item We investigate how accurately explanations can depict a detector's behavior \cite[\textbf{faithfulness:}][]{jacovi_towards_2020,ribeiro_why_2016,alvarez_melis_towards_2018}. This is crucial as unfaithful explanations can create a false sense of understanding.
    \item We study whether they are sensitive enough and sufficiently deterministic \cite[\textbf{stability}:][]{alvarez_melis_towards_2018,lakkaraju_robust_2020,nauta_anecdotal_2023}. Both aspects are prerequisites for building user trust.
    \item And we investigate how effective they are at communicating the model's decision process to users \cite[\textbf{usefulness}:][]{hoffman_metrics_2019,doshi-velez_towards_2017}, as explanations must be able to generate meaningful insights.
\end{enumerate}
Moreover, reliable explanation methods can generate insights into differences that exist between human- and generated language (e.g., syntactic structures, stylistic choices), while unreliable explanation methods may generate a false sense of understanding of those differences.
To assess how suitable these methods ultimately are for explaining detector decisions, we perform a systematic evaluation with automated metrics and a user study. We test for the above-mentioned aspects with two existing fine-tuned Transformer-based detectors \cite{guo_how_2023,solaiman_release_2019} and a zero-shot method \cite{mitchell_detectgpt_2023}.

To enable a comparison of faithfulness, we adapt the ideas of a \textit{token removal experiment} \cite[where the effect that removing tokens selected by the explanation method has on the classifier's decision is studied;][]{arras_explaining_2016} and the \textit{pointing game} \cite[Section \ref{methodfaitfulness};][]{poerner_evaluating_2019} to this task and all explanation methods, and we construct a dedicated test suite to measure the pointing game accuracy in this setting.
Stability is assessed with \textit{controlled synthetic data checks} \cite[which test if explanations are in line with a certain data generation strategy;][]{nauta_anecdotal_2023} that we construct specifically for the task of detecting MGT. 
Note that previous user studies have evaluated usefulness on tasks for which humans already have an intuitive understanding, such as sentiment analysis or income prediction \cite{ribeiro_why_2016,ribeiro_anchors_2018,hase_evaluating_2020}. 
Identifying patterns in a detector's behavior is, however, arguably \textbf{more challenging for users} when they have \textbf{limited knowledge about the features} that might be relevant. 
To keep \textit{forward simulation} experiments (in which users' ability to anticipate the classifier's behavior is measured before and after showing explanations) feasible for humans 
in a setting where detectors may utilize complex and unintuitive feature sets, we use a study design similar to that of \citet[][originally from the context of movie review sentiment prediction and income prediction]{hase_evaluating_2020} but add a special document selection strategy: Rather than randomly selecting documents, we present users with pairs of documents that we choose based on explanation similarity.

Figure \ref{fig1} presents the aggregated results of all our experiments. In summary, the contributions of our work are:
\begin{figure}[!ht]
\includegraphics[width=1\columnwidth]{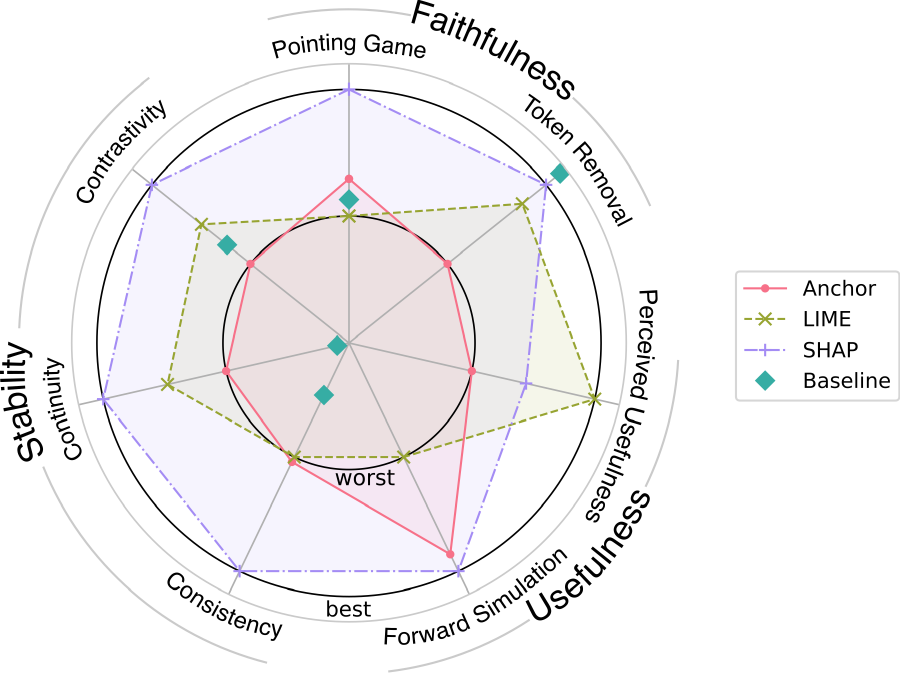}
\caption{{\bf Scores in the experiments.}
Min-max normalized for each respective metric.}
\label{fig1}
\end{figure}
\begin{enumerate}[nosep, topsep=0pt, label=(\arabic*), left=0pt]
    \item [(1)] The first study to systematically evaluate explanation methods for detectors of MGT, an important basis for understanding differences between generated and natural language; explanation methods can only be relied upon in critical scenarios if their strengths and limitations are experimentally validated.
    \item [(2)] Evaluation of all combinations of three detectors, three explanation methods with five automated experiments.
    \item [(3)] LM-assisted construction of task-specific test sets to measure contrastivity and faithfulness in this setting.
    \item [(4)] In addition to automated metrics, a user study to measure perceived- and actual usefulness in a forward simulation experiment.

\end{enumerate} 
 We find that SHAP performs best across all automated metrics. LIME explanations are perceived as most useful by the users but decrease users' performance at predicting the decisions of the detectors. Neither Anchor nor LIME consistently outranks the other in our experiments.\footnote{Code, data and explanations are made available at \href{https://doi.org/10.5281/zenodo.15655989}{doi.org/10.5281/zenodo.15655989}}
\section{Related work}\label{relateddwork}
\paragraph{Evaluating faithfulness. }
\citet{arras_explaining_2016} evaluate faithfulness \cite[the degree to which the explanation method accurately depicts the detector's behavior;][]{jacovi_towards_2020,ribeiro_why_2016,alvarez_melis_towards_2018} with token removal experiments. 
This type of experiment tests whether removing tokens found to be important by the explanation method is more likely to change the classifier's decision than removing random tokens.
Note that this approach raises the same concern that perturbation-based explanation methods do when applied to text classification:
It assumes that classifiers behave predictably for partial input, which is problematic for the task at hand as (accidental) omission of words might be an effective feature for distinguishing human-written from machine-generated text.
\textit{Controlled synthetic data checks}~\cite{nauta_anecdotal_2023} are an alternative type of experiment that foregoes this issue. \citet{poerner_evaluating_2019} propose a design of this type for evaluating local feature importance explanations for sentiment prediction. We extend both experiments to the task of MGT detection, including rule-based Anchor explanations, in Section \ref{methodfaitfulness}.

\paragraph{Evaluating stability. }
Quantifying the stability of explanations across runs with agreement measures (\textit{consistency}) is discussed in \citet{nauta_anecdotal_2023}.
The properties of \textit{continuity} \cite[similar explanations for similar documents with the same prediction;][]{alvarez_melis_towards_2018} and \textit{contrastivity} (sufficiently different explanations for similar documents but different predictions) described by \citet{nauta_anecdotal_2023} are alternative notions of stability that rely on convergence to a lesser extent.
We propose a setup that generates coherent perturbations from a language model for these tests in Section \ref{continuity}.

\paragraph{Evaluating usefulness. }
\citet{hase_evaluating_2020} perform a \textit{forward simulation} experiment \cite{doshi-velez_towards_2017}, 
where they assess whether users can predict the detector's behavior better after being shown explanations.
\citeauthor{hase_evaluating_2020} conduct their user study with two comparatively simple binary classifiers and tasks (sentiment analysis and income prediction). 
Given that the explanation methods analyzed here produce outcome- and not model explanations \cite[i.e., they explain one instance at a time;][]{guidotti_survey_2018}, it is less likely that users will obtain a comprehensive understanding of the model's behavior or relevant features and successfully apply that to new instances of human-written and MGT.
We propose a strategy to increase the feasibility of this kind of experiment in Section \ref{setup-user-study}.
Besides measuring user performance, we also assess \textit{perceived usefulness}. This is measured in a \textit{rating task} with questions adapted from \citet{hoffman_metrics_2019}.

\section{Methods: Explanation Quality Metrics for MGT Detectors}\label{methodology_eval_methods}
We systematically evaluate explanation quality along the axes of faithfulness, stability, and usefulness.
Throughout this section, $f(d_i)$ refers to the decision of a detector $f$ for a document $d_i$. $D$ is the base dataset of human-written and machine-generated documents $\{d_i\}$. 
SHAP and LIME provide \textit{feature importance scores} for tokens from the input document $d_i$ as explanations for decisions made by $f$~\cite{lundberg_unified_2017}.
Those tokens that had the strongest impact on the detector's decision should be attributed the highest scores. 
SHAP- and LIME explanations are largely similar from a user's perspective, as they share the same explanation format and rely on similar visualization methods (e.g., feature importance plots: Figure \ref{fig4} and Appendix \ref{S1_Appendix}).
We describe how these two methods differ in their calculation of feature importance in Section \ref{methodexplanationmethods}.

Anchor explanations are expressed as \textit{if-then rules}. \citet{ribeiro_anchors_2018} define an Anchor $A_{i,m}$ for the document $d_i$ as a set of tokens $A_{i,m} = \{t_{i,j}\}$ that, if present in the document, guarantee the same decision as for $d_i$ with a probability greater than $\tau$ for perturbations in the local neighborhood of $d_i$. There can be multiple valid Anchors per document.

\subsection{Evaluation of Faithfulness}\label{methodfaitfulness}
A faithful explanation method should accurately depict the detector's behavior~\cite{jacovi_towards_2020,ribeiro_why_2016,alvarez_melis_towards_2018}.
The first test for this is a \textbf{pointing game} similar to that in \citet{poerner_evaluating_2019}. We first describe the approach for feature importance explanations (SHAP and LIME) before extending this metric to rule-based explanations (Anchor).

\begin{figure}[!ht]
\includegraphics[
    width=1\columnwidth,
    trim=0.6cm 0.5cm 1cm 0.3cm,  
    clip
]{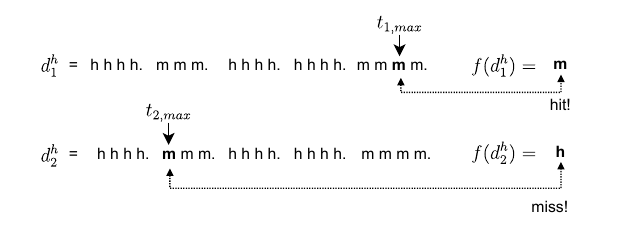}

\caption{{\bf Hybrid documents in the pointing game.}
The detector's prediction for a hybrid document $f(d^h_i)$ is compared against the ground truth of $t_{i,max}$ (the token with the highest importance score in the explanation). Sentences labeled with \textit{m} or \textit{h} originate from machine-generated or human-written documents respectively}
\label{fig2}
\end{figure}

Random sentences from the base dataset $D$ are concatenated to form a synthetic dataset of \textit{hybrid documents} $D^h = \{d_{i}^{h}\}$ (Figure \ref{fig2}). In our setup, we set the number of sentences per hybrid document to the mean number of sentences per document in the original dataset, and sample random sentences without replacement until all of $D$ is used.

\noindent It is assumed that the detector's decision $f(d_{i}^h)$ on a hybrid document  
is based on segments that were originally part of documents with ground truth $y_i$ equal to $ 
 f(d_{i}^h)$.
A faithful explanation method should hence find these segments to be more important for the decision than those originating from opposite-class documents.
In the pointing game, feature importance explanation methods are therefore awarded hits for a document $d_i^h$ if the token with the highest feature importance score $t_{i,max} \in d_i^h$ 
was in fact originally part of a document with ground truth $y_i$ equal to $f(d_{i}^h)$~\cite{poerner_evaluating_2019}: 
\begin{equation}
\begin{split}
    S_x = \bigcup_{\forall d_i \in D}\{t \in d_i \,\mid\, y_i = x\},\\
    hit(d_i^h,t) = \mathbf{1}[t \in S_{f(d_i^h)}]
\end{split}
\end{equation}
Where $x$ is a label (\textit{human} or \textit{machine}) predicted for a particular hybrid document so that $S_{human}$ is the set of all tokens from documents in $D$ with ground-truth $human$, and $\mathbf{1}[\cdot]$ is 1 if $\cdot$ evaluates to true, 0 otherwise.
The \textit{pointing game accuracy} is the fraction of documents in the dataset $D^h$ that get awarded hits. For feature importance type explanations, it is given as: 
\begin{equation}
    \text{Acc}_\text{pg} = \sum\limits_{\forall d_i^h \in D^h}{hit(d_i^h, t_{i,max})} / |D^h|
\end{equation}

\noindent \citet{poerner_evaluating_2019} only evaluate feature-importance explanation methods. We add support for rule-based explanations as follows:
First, there is no distinction by importance between tokens within an Anchor. 
Furthermore, a single Anchor can span multiple sentences. We therefore attribute hits proportionally for this explanation method. 
The hit function then tests all tokens $t_{i,j}$ specified by the Anchor $A_i$ individually and returns the average number of hits instead of a binary value:
\begin{equation}
    hit^{R}(d_i^h)  =\frac{\sum_{\forall t_{i,j} \in A_i} hit(d_i^h, t_{i,j})} {|A_i|}
\end{equation}

We perform a \textbf{token removal experiment} as in \citet{arras_explaining_2016} as the second test for faithfulness. 
The assumption in this experiment is that the prediction is more likely to change if tokens important to the detector's decision are removed, rather than random ones.
For the two feature importance explanation methods (SHAP and LIME), let $d_i$ be the original document and $d_i^{k}$ a version with the $k$ top-tokens by feature importance towards $f(d_i)$ removed. \citeauthor{arras_explaining_2016} then plot the accuracy of the detector at different $k$ with respect to the ground truth $y_i$ of the original documents. 

While rule-based Anchor explanations do not rank tokens by their importance, they still aim to select tokens that were highly important for the detector's decision. 
To emulate the logic outlined above for this form of explanation, we remove tokens from the Anchor that applies to the highest proportion of documents in the local neighborhood (the one with the highest $\tau$) in random order.
We perform token removal at increments of 1 for $k \in [0,10]$, at $k=15$, and then increments of 10 up to $k=100$.

\subsection{Evaluation of Stability}\label{methodstability}
As discussed in Section \ref{relateddwork}, stability has several definitions in the existing literature. We therefore propose metrics for three different notions of stability. Specifically, our experiments aim to assess the properties of consistency, continuity, and contrastivity as characterized in ~\citet{nauta_anecdotal_2023}. 
For simplicity, we describe these properties and our experiments for feature-importance explanation methods here. Anchor explanations are transformed into feature-importance explanations using one-hot encoding: if a token in a document is part of an Anchor, it is considered important.

\paragraph{Consistency.} We measure consistency (stability of explanations between runs) across five explanations for the same document with the agreement metric Krippendorff's $\alpha$~\cite{krippendorff_bivariate_1970}, calculated on the explanations' feature-importance vectors.

\paragraph{Continuity.} \label{continuity}
We measure continuity \cite[similar explanations for similar documents with the same prediction;][]{alvarez_melis_towards_2018} between the explanation for the original document and explanations for a set of 5 perturbations $\{d_{i}^{\iota}\}$ with Krippendorff's $\alpha$ as well. 
For each $d_i$ from the original dataset, a single token is randomly selected and replaced with an arbitrary number of tokens using the T5 language model (\verb|t5-small| 60.5M params: \citealp{raffel_exploring_2020}). It is verified that $f(d_i^{\iota}) = f(d_i)$ for all 5 perturbations $d_i^{\iota}$. We rely on T5 for generating synthetic documents because replacement with random tokens makes text appear more human-written to the detectors (likely due to decreased coherence). This makes the construction of valid perturbations through random replacement for machine-generated documents within a reasonable amount of attempts infeasible.
While this adds machine-generated parts to all perturbations, we are only interested in and only calculate Krippendorff's $\alpha$ on the matching parts of the original and perturbed documents.  Note that we do not test for faithfulness in this experiment. This still aligns with the rationale behind the property of continuity, which states that the rest of the explanation should not drastically change if the classifier's prediction remains the same \cite[][regardless of the type of signal introduced]{nauta_anecdotal_2023}. 
We set the maximum output length to 150 tokens.  
In some instances, T5 fails to generate five unique replacements. In these cases, the token is replaced with a random token from the vocabulary.

\begin{figure*}[!ht]
\centering
\includegraphics[width=0.9\textwidth]{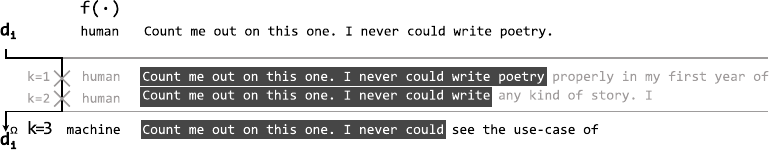}
\caption{{\bf Perturbation strategy for the contrastivity experiment.}}
\label{fig3}
\end{figure*}

\paragraph{Contrastivity. } 
The high-level goal of contrastivity evaluation is to verify that documents that are similar in content, but are assigned different labels by the detector, get sufficiently different explanations~\cite{nauta_anecdotal_2023}.
We implement this principle by editing documents from the dataset with a language model in order to obtain pairs of coherent documents $(d_i,d_i^{\Omega})$ where $d_i$ and $d_i^{\Omega}$ are similar but get assigned different labels. 
In each iteration $k$, one token is deleted from the end of the document (Figure \ref{fig3}). The shortened document $d_i^*$ is used as an initial prefix for the autoregressive foundation model \verb|facebook/opt-350m|: \cite{zhang_opt_2022} to generate a continuation that (together with the prefix) will constitute the perturbation $d_i^{\Omega}$. We halt removing tokens as soon as $f(d_i^{\Omega}) \neq f(d_i)$.  

To increase the chance of a label flip with smaller edits, five attempts are made at every $k$.
Still, if the first tokens in the document are highly influential for the decision of the detector, more tokens need to be deleted until a label change can be observed, resulting in more dissimilar perturbed documents (e.g., documents that start with greetings like "Hi!").
To retain a certain level of similarity between documents, perturbations that edit more than 50\% of the original document, are discarded and not used for calculating the metric. The resulting synthetic datasets of pairs (one set per detector) are further described in Appendix \ref{S2_Appendix}. 

Two scores are subsequently calculated on the explanations for $d_i$ and $d_i^{\Omega}$. These verify whether the explanations are consistent with the generation strategy: 
Given that the left part of $d_i$ and $d_i^{\Omega}$ are identical, but $f(d_i^{\Omega}) \neq f(d_i)$, one expects the filled-in part in $d_i^{\Omega}$ to have had a strong influence on the detector when labeling $d_i^{\Omega}$. A sufficiently sensitive explanation method should detect this.

The scores are formulated as synthetic data checks, similar to the the pointing game described above. 
Note that unlike with that experiment, but similar to the experiment testing for continuity, it is sufficient here that the modification we apply to a given document causes the detector to predict the opposite label. This is because we only aim to verify whether the explanations exhibit a desired behavior (that tey appear sufficiently different if the documents are assigned different labels, not whether they are faithful), which does not require that the signal we introduce is of the opposite type.

The first score, \textbf{$c_\textit{inter}$} is calculated on the parts that differ across the two documents, $d^{-}_i$ and $d^{\Omega-}_i$. It tests whether the mean feature importance score towards $f(d^{\Omega}_i)$ (denoted $\mu_{\vec{v}^{~\Omega}}(\cdot)$) is higher for $d_i^{\Omega-}$ than it is for $d_i^{-}$. The hit function for this case is: 
\begin{equation}
hit_\textit{inter} = \mathbf{1}[\mu_{\vec{v}^{~\Omega}}(d_i^{\Omega-}) > \mu_{\vec{v}^{~\Omega}}(d_i^-)]
\end{equation}

\paragraph{$c_\textit{intra}$} shares the same intuition but is defined only on $d_i^{\Omega}$. As the left parts of $d_i$ and $d_i^{\Omega}$ are identical, but $f(d^{\Omega}) \neq f(d_i)$, one expects the generated part $d_i^{\Omega-}$ to have a higher average feature importance score towards $f(d_i^\Omega)$ than the shared part: 

\begin{equation}
    hit_\textit{intra} = \mathbf{1}[\mu_{\vec{v}^{~\Omega}}(d_i^{\Omega-}) > \mu_{\vec{v}^{~\Omega}}(d_i^{\Omega*})]
\end{equation}
\noindent Both scores are again given as the fraction of documents in the dataset that score a hit: 
\begin{equation}
    c_{\{\textit{intra},\textit{inter}\}} = \sum\limits_{\forall (d_i,d_i^{\Omega})}{hit_{\{\textit{intra},\textit{inter}\}}(d_i,d_i^{\Omega})} / |D|\notag
\end{equation}
\subsection{Evaluation of Usefulness}\label{setup-user-study}

\begin{figure*}[!ht]
\centering
\includegraphics[width=0.9\textwidth]{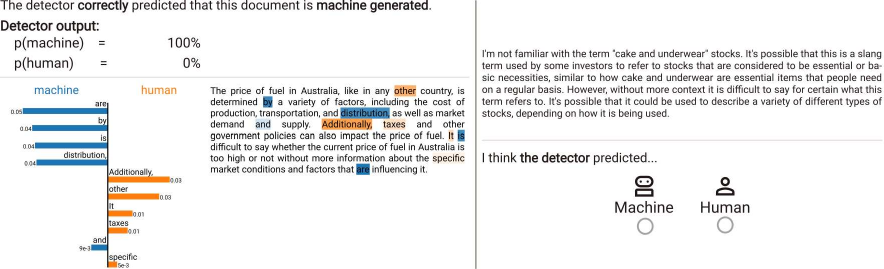}
\caption{{\bf User Study.} Information shown for LIME in phase 3 (left); and in the annotation phases 2 and 4 (right). }
\label{fig4}
\end{figure*}

We define \textit{usefulness} as the explanation method's ability to improve users' understanding of the detector's behavior. 
Note that we do not aim to measure users' performance at the base task of detecting machine-generated text. Instead, our goal is to measure their ability to predict the detector's decision, both with and without explanations.
To do so, we modify the design for a forward simulation experiment in \citet{hase_evaluating_2020}.
As in their design, in phase 1 of the experiment, users inspect decisions of the detector on a set of documents $\{a_j\}$ without explanations. In phase 2 they are then instructed to anticipate the detector's decision (\underline{not} to guess the true document class) on a second set $\{b_j\}$. Phase 3 provides explanations for set $\{a_j\}$ (Figure \ref{fig4}). Users conclude the experiment by labeling the documents from set $\{b_j\}$ again in phase 4. 
The change in user accuracy from phase 2 to 4 is reported as a measure of performance. We randomize example ordering between participants.

While relying on this single performance measure makes it difficult to isolate the direct impact of explanations from users' growing familiarity with features indicative of human-written or machine-generated text, our setup is designed to reflect typical real-world model inspection: In such contexts, users need to examine numerous explanations to adequately learn about a classifier. Crucially, this design still allows for a fair \textit{relative} comparison of different explanation methods.

We add three additional questions from \citet{hoffman_metrics_2019}, asked for every explanation shown in phase 3 to measure \textit{perceived usefulness}. These are provided in Appendix \ref{S1_Appendix}. In order to be able to track whether users can pick up regularities between the different phases, we apply a document selection strategy that ensures a minimal overlap of features in the documents, see Appendix \ref{S1_Appendix}. 
\section{Technical and Experimental Details}\label{td}
\subsection{Dataset}\label{dataset}
We use a subset of the H3 dataset by \citet{guo_how_2023} for our experiments. They add ChatGPT written answers to question answering datasets from various domains (ELI5: \citealp{fan_eli5_2019}, WikiQA: \citealp{yang_wikiqa_2015}, FiQA: \citealp{maia_www18_2018}, Medical Dialog: \citealp{he_meddialog_2020} and Wikipedia computer science articles: \citealp{guo_how_2023}; published under the CC-BY-SA license).
We remove all questions from WikiQA and ELI5, as the source datasets contain crawling artifacts that make identifying human texts trivial.  
Moreover, only documents that are between 50 and 150 words long are kept. This is to ensure a sufficient input length for the zero-shot detector. In the end, 1016 documents remain. 
Due to the high computation time involved with generating explanations, only 30\% of documents (N=305) are used (stratified split of machine-generated and human-written documents) for our experiments. As discussed in Section \ref{methodology_eval_methods}, however, we derive additional synthetic datasets from this set, and repeat all experiments with 3 detectors. 
\subsection{Detectors}\label{methoddetectors}
We apply the explanation methods to three \textbf{detectors} of machine-generated text:


\noindent

  \paragraph{Guo \cite{guo_how_2023}.} With their dataset, \citeauthor{guo_how_2023} also ship a fine-tuned RoBERTa model (\verb|roberta-base| 125M: \citealp{liu_roberta_2019}) . The accuracy of this detector on the sub-set for which explanations were generated is 0.99.
  \paragraph{Solaiman \cite{solaiman_release_2019}.} This detector was fine-tuned on the Webtext and GPT-2 output datasets \cite{solaiman_release_2019}, but uses the same base model as \citet{guo_how_2023}. It is therefore treated as an out-of-distribution detector. Its accuracy is 0.92. 
  \paragraph{DetectGPT \cite{mitchell_detectgpt_2023}.} This is a zero-shot method. It is set up here with a smaller language model (\verb|pythia-70m|: \citealp{biderman_pythia_2023}) as suggested by  \citet{mireshghallah_smaller_2023} and accomplishes an accuracy of 0.74. This, and reducing the number of perturbations per evaluation from 100 to 5 is done to reduce inference time from roughly 6.3 (15.8 with GPT-2) to 0.9 seconds per document. The default setting for this detector would require $10^5$ generations with GPT-2 for a single explanation. A comparison with the original implementation is provided in Appendix \ref{S3_Appendix}.

\subsection{Explanation Methods}\label{methodexplanationmethods}
\paragraph{Feature importance explanation methods}
assign scores to individual tokens in the document with the goal to quantify the effect a token has on the decision of the classifier. 
\textbf{LIME} trains a local surrogate model on a set of data point perturbations and corresponding detector outputs \cite{ribeiro_why_2016}. 
The number of perturbations to use for fitting this model, and its size, have to be chosen manually. 
The number of samples was set to 1k  
for the RoBERTa-based detectors and 500 for the zero-shot method to match SHAP's runtime. This results in an average of $\approx$30s for the RoBERTa-based detectors and $\approx$415s for DetectGPT per explanation.
The default number of 10 features to show as an explanation (Figure \ref{fig4}) appears appropriate for the document length used here (50-150 tokens).

The implementation of \textbf{SHAP} we use (Partition Explainer; \citealp{lundberg_unified_2017}; default parameters) computes Owen values \cite{owen_owen_1977} as a measure of feature importance. 

\paragraph{Rule-based explanation methods} -- like \textbf{Anchor}, explain complex decision processes with short rules \cite{ribeiro_anchors_2018}.
Among valid rules as outlined in Section \ref{methodology_eval_methods}, Anchor's search algorithm attempts to select those that apply to the highest proportion of perturbations in the local neighborhood. In its default implementation, Anchor is often unable to terminate for a single document within an hour of runtime with these detectors. We employ the following strategies to make the computation of Anchors feasible here:
We choose a low target level of precision ($\tau = 0.75$) and impose a limit on the number of samples used during construction (200 samples per candidate Anchor).
For generating perturbations, we employ DistillRoBERTa (\verb|distilroberta-base|: 82.8M params \citealp{sanh_distilbert_2020}) instead of DistillBERT, to increase their coherence and edit at most 20\% of tokens per perturbation. 
Combined, this reduces the computation time to an average of roughly 5 minutes per explanation for the RoBERTa-based detectors and 15 minutes for DetectGPT on the original documents.

\paragraph{Perturbation strategy.}
\noindent An experiment that tests the effect of different perturbation strategies (Appendix \ref{S2_Appendix}) did not single out a method that works equally well for all detectors.
In the interest of consistency, perturbations are generated by replacing tokens with the mask token of the detector's tokenizer for both LIME and SHAP throughout all experiments. 
Anchor offers masking with a specified token or to perturb with a language model. The latter strategy was used, as it was found to terminate considerably faster with these detectors.

\subsection{User Study}\label{userpooluserstudy}
We recruited 36 participants (B.Sc., M.Sc., and PhD students) with a background in computer science and English reading proficiency at the C1 CEFR level or higher between 20/03/2024--12/04/2024. Of them, 27 stated that they had never worked with explanation methods before. The instructions shown to users on the individual explanation methods are provided in Appendix \ref{S1_Appendix}.
The participants were offered a compensation of €10 for their participation. The experiment was conducted online through a purpose-built web service, through which we also documented informed consent.
We use the default presentation formats of the respective explanation methods to ensure that our analysis is informative about existing applications on MGT detection  \cite[e.g.,][]{mosca_distinguishing_2023, liu_check_2023, yu_cheat_2023}.

\section{Results}\label{resultsmain}
\begin{table*}[ht] 
\centering
\caption{{\bf Aggregate results for the faithfulness and stability experiments.} Mean scores across the three detectors used for evaluation. The best scores are highlighted in \textbf{bold}.}
\resizebox{0.8\textwidth}{!}{\begin{tabular}{c|c|cc|c|c|ccc}
\hline
 & \textbf{Pointing Game} & \multicolumn{2}{c|}{\textbf{Token Removal}}&  \textbf{Consistency}  &  \textbf{Continuity}     & \multicolumn{3}{c}{\textbf{Contrastivity}} \\
 &  $\text{Acc}_{pg}$  &
 $\Delta_{\text{right},k=10}$ & $\Delta_{\text{wrong},k=10}$& 
 $\alpha$  & $\alpha$    & $c_\textit{inter}$ & $c_\textit{intra}$  \\

\hline
Random\ &0.565& 51.2\%& 75.1\%   & -0.167   & -0.139 & 0.486 & 0.498 \\
\hline
LIME\ &0.546 & 46.9\% & 57.3\%   & 0.136  & 0.394  & 0.598 & 0.587\\
\hline

Anchor\ & 0.589 & 28.3\% & \textbf{100.0}\%  & 0.160 & 0.210 & 0.512 & 0.289\\
\hline

SHAP\ & \bfseries 0.692 & \textbf{50.4}\% & 62.3\%    & \bfseries 0.695 & \bfseries 0.596  &  \textbf{0.799} &  \textbf{0.774}  \\
\hline

\end{tabular}}

\label{tab:main_table}
\end{table*}

\noindent Aggregate results per explanation method for faithfulness and stability are reported in Table \ref{tab:main_table}, those from the user study in Table \ref{results_user_study_detector_combined}. See Appendix \ref{S2_Appendix} for results per detector-explanation method pairing. We provide the mean scores for five explanations consisting of random feature importance vectors as a baseline.
\subsection{Faithfulness}
In the \textbf{pointing game} (Table \ref{tab:main_table}), SHAP performs best and a series of binomial tests ($H_0:$ No difference between one method and the next best in the ranking)  
shows that the difference between SHAP and Anchor, as well as the one between Anchor and the random baseline are significant ($p < 0.05$). LIME does not outperform a baseline of random feature importance scores but is also not significantly worse ($p=0.143$).

\begin{figure*}[ht]
\includegraphics[width=1\textwidth]{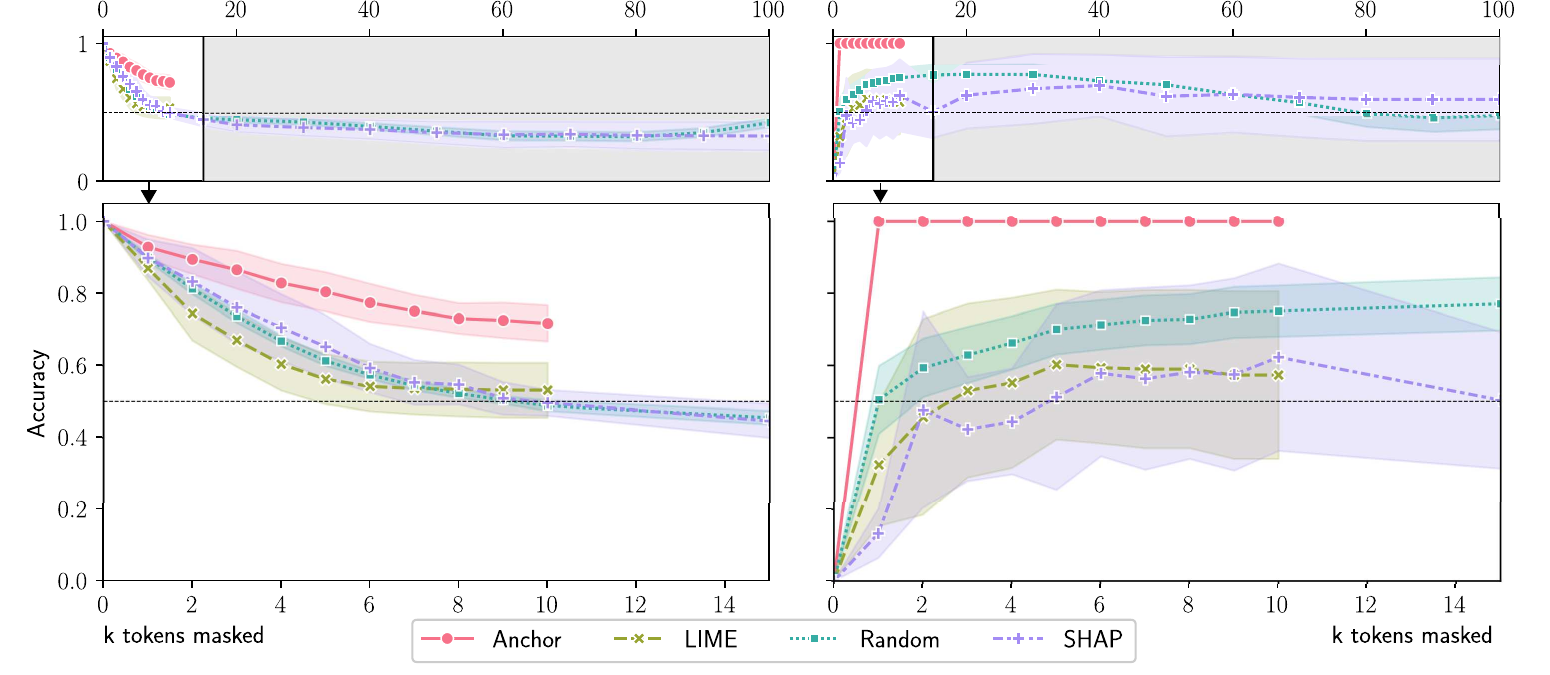}
\caption{{\bf Accuracy at $k$ tokens masked.}
A faithful explanation method should feature a steep decline (initially correct predictions, left) or steep incline (initially wrong predictions, right).  Only SHAP explanations cover more than 10 tokens. Mean across all detectors, error bars at $\pm 1$ standard error.}
\label{fig5}
\end{figure*}

For the \textbf{token removal experiment}, 
results for initially correct predictions ($f(d_i) = y_i$) and initially wrong predictions ($f(d_i) \neq y_i$) are plotted separately in Figure \ref{fig5} to allow for consistent interpretation of accuracy scores \cite{arras_explaining_2016}. We report the change in accuracy at $k=10$ tokens removed ($\Delta_{k=10}$) in Table \ref{tab:main_table}, corresponding to the maximum number of tokens Anchor and LIME include in their explanations.
For $f(d_i) = y_i$, the most important features are removed first. A faithful (feature importance) explanation method should show a steep drop in accuracy. For $k < 10$ tokens removed, the accuracy of LIME drops slightly faster than that of SHAP. The average accuracy drops below 50\% for SHAP at roughly 10 tokens masked. For Anchor, the accuracy drops slower than for all other methods.

For $f(d_i) \neq y_i$, one expects the accuracy to increase. Note that few initially wrong examples are available (105 vs. 810 initially correct cases) given the high accuracy of two detectors. 
The average accuracy of LIME and SHAP does not increase faster than the random baseline here. Anchor flips the label in all instances and thus archives a perfect score at $k=10$ for initially wrong examples.
\subsection{Stability}
\paragraph{Consistency. } 
For Krippendorff's $\alpha $, a score of 0 reflects an agreement by chance, 1 perfect agreement, and negative values systematic disagreement \cite{krippendorff_bivariate_1970}.
Scoring 0.136 and 0.160 respectively, LIME's and Anchor's consistency in re-runs is far below what could be considered reliable. SHAP (Partition Explainer) is deterministic for the detectors of \citet{guo_how_2023} and \citet{solaiman_release_2019}. Note that the zero-shot detector of \citet{mitchell_detectgpt_2023} is not deterministic. We include this detector in the aggregate scores in Table \ref{tab:main_table} for consistency with the other experiments. The ranking remains unaffected by this, per-detector results are provided in Appendix \ref{S2_Appendix}.

\paragraph{Continuity. } 
SHAP's explanations remain stable under small perturbations ($\alpha > 0.8$ for the detectors of \citealp{guo_how_2023} and \citealp{solaiman_release_2019}). Those from LIME and Anchor do not, but agree with each other better than explanations from the random baseline.  

\paragraph{Contrastivity.}
SHAP ranks highest in both synthetic data checks $c_\text{inter}$ and $c_\text{intra}$: It can identify that the filled-in part is responsible for flipping the label more reliably than LIME and Anchor.

\subsection{Usefulness}\label{resultsusefulness}
\begin{table*}[ht!]
\centering
\footnotesize
\caption{{\bf Results from the user study.} 5-point Likert scale (3 = neutral, 5 = strongly agree).}
\begin{tabular}{l|rrrr|rrr} \hline
&\multicolumn{4}{c|}{Forward Simulation} & \multicolumn{3}{c}{Perceived Usefulness}\\
 & Without & With & Change & \multicolumn{1}{c|}{p} & Q1: \textit{Why} & Q2: \textit{How} & Q3: \textit{Helpful}\\ \hline
LIME & 0.741 & 0.644 & -13.12\% & 0.006 & \textbf{3.60} & \textbf{3.37} & \textbf{3.31}\\ 
Anchor & 0.694 & 0.699 & 0.67\% & 1.000 & 2.57 &2.48 & 2.51\\
SHAP & 0.755 & 0.778 & \textbf{3.07\%} & 0.551 & 3.06 & 2.86 & 2.84\\\hline
\end{tabular}

\label{results_user_study_detector_combined}
\end{table*}
\paragraph{Forward simulation. }
Table \ref{results_user_study_detector_combined} shows the change in user accuracy from phase 2 to phase 4 and the results from McNemar's tests. SHAP is the best-performing method, followed by Anchor and LIME. The increase in user accuracy for both SHAP and Anchor is not significant at $p < 0.05$. 
However, participants who have seen LIME explanations perform 13.12\% worse after being shown explanations than they did before (p=0.006).

\paragraph{Perceived usefulness. }
Conversely, users rated LIME best, SHAP second, and Anchor third across all three questions assessing perceived usefulness. For SHAP, users only tended to agree with the first item, where they were asked whether they could \textit{understand why} the detector decided the way it did from an explanation (annotator instructions provided in Appendix \ref{S1_Appendix}).
They disagreed with the two other items, asking whether they better \textit{understood how} the detector works and whether they thought the information from the explanation \textit{would help them} perform better in the second round of annotation. 
They tended to agree for LIME with all three statements, for Anchor to disagree  on average.

\section{Discussion}\label{discussion}
\subsection{Faithfulness} 
The explanations from SHAP and Anchor are in line with the data generation strategy in the \textbf{pointing game}, and LIME achieves the lowest pointing game accuracies. 
The pointing game and the token removal experiment are based on different assumptions: The pointing game rewards if the explanation method is consistent with the assumption that a classifier relies on material that is associated with a predicted class more than with unrelated material. This is a reasonable expectation, especially for classifiers with much better than random accuracy. The \textbf{token removal experiment} rewards if the explanation method ranks features on top that if removed in a greedy fashion influence prediction the most. This \textbf{might be at odds} with other notions of importance, such as Shapley values, that estimate importance averaged over different combinations of features. Furthermore, it remains unclear whether a steeper initial drop in accuracy, as observed for LIME, or an overall higher drop in accuracy, as observed for SHAP, is more desirable (see left plot in Figure \ref{fig5}), especially given that explanation length is ultimately a design choice: LIME explanations show only the 10 most influential words by default, whereas SHAP provides scores for every word in the input in our setup.

The definition of an Anchor states that changing tokens not part of it should not affect the detector's prediction.
 Therefore, one would expect that masking tokens which are part of an Anchor would change the prediction more frequently than random masking.
One should, however, not expect the change in accuracy to be more pronounced than for the other methods:
Anchor does not attempt to identify a set of tokens that affect the decision most. 
It merely aims to provide a set that is \textit{important enough} to cause a certain outcome. 
While Anchor is less accurate in the token removal experiment for the majority of cases (where the detectors predict correctly), it however outperforms the other methods for the much smaller class of initially wrong predictions.

\subsection{Stability}
The partition tree used by SHAP's Partition Explainer is obtained with a deterministic algorithm. SHAP is therefore able to accomplish a perfect score in the experiment for \textbf{consistency} (across re-runs) for the two deterministic detectors (in contrast to LIME that uses a different random seed and set of perturbations for each explanation attempt). 
This is precisely why we included two additional measures of stability that rely on convergence to a lesser extent: in fact, the difference between SHAP, LIME, and Anchor is \textbf{less pronounced} in the \textbf{continuity} experiment, which measures how explanations are affected by small changes to the input documents. 
Here, the assumption is that similar documents -- from the perspective of the detector -- should have similar explanations.
We use a \textbf{two step-process} to obtain similar perturbations: first, we obtain \emph{semantically} similar documents by querying language models to provide partial reformulation. In contrast to explanations for vision tasks (e.g., \citealp{alvarez_melis_towards_2018}; where \emph{continuous} perturbations can be chosen such that they have a limited impact on the classifier output to be explained), perturbation in discrete input domains such as text can have more severe impacts on the prediction.
We limit this unwanted effect by discarding perturbations that would flip the label prediction.
However, filtering by label flip is \textbf{just a proxy for similarity} from the detector's perspective, and does not differentiate between different types of effects that could lead to the same predicted label:
some reformulations may feature low similarity to the original text (and necessitate dissimilar explanations to retain faithfulness), but be assigned the same label by the detector.
Consequently, our continuity experiments \textbf{may underestimate the agreement of explanations} (high agreement equals high continuity).

Regarding \textbf{contrastivity}, where we assess whether explanations for similar documents with different predictions are sufficiently different, SHAP is more performant than LIME and Anchor: SHAP attributes importance in the expected way more often than the other methods ($c_\text{inter}$).
SHAP is also considerably more successful at identifying that the filled-in part, and not the shared part, is responsible for flipping the label ($c_\text{intra}$). 

Based on the results from these three experiments, \textbf{SHAP} appears to be the \textbf{most sensitive} method and to produce \textbf{more stable} explanations than LIME and Anchor in our setup. 

\subsection{Usefulness}
When measuring the usefulness of explanations presented to users, different aspects must be differentiated: a (partial) \textbf{understanding} of the model (the detector in our case) vs. the \textbf{perceived degree of understanding} of the model. These two may not be the same, it can be that an explanation model gives plausible explanations that do not provide actual insight into the detector's behaviour. Another subtle distinction is that between (perceived or actual) understanding of model behaviour \textbf{vs. understanding of task characteristics} (what makes human and generated texts different).
We have tested for model understanding in the forward simulation experiment, where we measure the effect of seeing predictions without and with explanations on the ability to anticipate model behavior on new inputs. 
We also reported \emph{perceived} usefulness of the different explanations.
However, even though we carefully phrased the questionnaire to inquire the perceived ability to understand model behaviour (\emph{I now better understand how the detector works}), it is conceivable that \textbf{users conflated} this with better understanding of task characteristics.

Our results show that none of the methods substantially increased the users' ability to anticipate detector predictions (with SHAP and Anchor having a small positive but not statistically significant effect).
One reason could be that in contrast to simpler tasks for which feature-based explanation methods had been applied to before (e.g., binary sentiment prediction), the possible feature space is much larger for the MGT detection task (obvious sentiment-bearing words vs. subtle frequency choices or constructions).
Good performance in the automated metrics ( faithfulness and stability) \textbf{did not translate to usefulness in the forward simulation}. 
The method with the best perceived usefulness (LIME) shows the worst performance there (as well as in many of the automated metrics).
We hypothesize that this might be partially due to differences in explanation length: LIME explanations are limited to the 10 most influential words (see Figure \ref{fig4}) whereas SHAP provides importance scores for every word in the input. Given the between-subject setup of our user study, an analysis of this factor will need to be explored in future work.
Our study is a strong motivation to rethink the applicability of local feature-based explanations to complex tasks, and highlights the importance of accompanying automated evaluation of explanation methods with user studies.

\section{Conclusion}\label{conclusion}
In this work, we conduct the first evaluation of explanation methods for detectors of machine-generated text.
We find that SHAP fulfills the theoretically motivated properties of good explanations of faithfulness, stability, and usefulness best in our experiments. However, no explanation method led to a significant increase in performance in a user study that tested users' ability to predict the behavior of detectors with the help of explanations.
Notably, for LIME, users' perceived usefulness of explanations did not align with the measured decrease in user's performance, nor with the results from the faithfulness and stability experiments. 
Given the results from the user study, we advise against implementing combinations of these methods and detectors in their current form into systems that face untrained users. 
We do however see their exploratory value for model- or dataset inspection. Based on the results from the faithfulness and stability experiments, we recommend SHAP for this application.

  We have focused on providing a relative comparison between explanation methods, and gathered insights into the suitability of feature importance explanations for tasks with complex feature sets. We did not however, aim to provide insights into specific detectors or to document useful features for MGT detection itself, as that would have required a dedicated analysis of individual explanation method- and detectors pairings, which is beyond the scope of our this work.

Ultimately, our study serves to demonstrate the necessity of evaluating model-agnostic explanation methods before applying them to new tasks and models. 
While our study can only partially indicate the general suitability of feature importance as an explanation format for language-model-based classifiers, it points to the need for further evaluation: specifically, questions remain about when feature importance provides sufficient insights and for which tasks it is most effective. 
We advocate for evaluation across additional tasks to better understand the conditions under which feature importance can be an adequate explanation format.

\section*{Limitations}\label{limitations}
We restrict our study to explanation methods that can be applied to black box predictors of machine-generated text, i.e., methods that generate explanations by tracing causal effects from manipulating inputs and observing the corresponding predictions. These methods are generally applicable to all detectors of generated text, and do not have access to training data, activations, attention patterns, and other internal states or detector-specific information. SHAP and LIME, the most prominent explanation methods for classifiers, fall into this category. Such black-box explanations allow for comparisons across detectors, because they do not depend on detector-specific properties. 
On the flip-side, explanation methods for black-box predictors rely on input-perturbations, and model-specific methods (such as attention patterns) or explanations that trace back training-data influence are not included in our study. 

More generally, most current explanation methods only provide explanations by identifying prediction regularities or causal structures of the predictor, but do not build a model of the human users to which the explanations are shown, and therefore may not be considered full explanations according to the theory of mind (i.e., they do not take into account what can be expected to be already known by a user). However, we argue that faithful, stable and useful feature-importance explanations will be an important building block for future explanation setups that combine such explanation algorithms with an elaborate model of the system context and a tailored user interface, and which will include a theory of mind of individual system users.

We restrict our study to three representative language-model based detectors and one popular benchmark dataset, due to resource constraints imposed by the inclusion of a user study.

\section*{Acknowledgments}
This research has been funded by the Vienna Science and Technology Fund (WWTF)[10.47379/VRG19008] "Knowledge-infused Deep Learning for Natural Language Processing".

\bibliography{main}

\appendix
\section{User study}\label{S1_Appendix}
\paragraph{Document selection.}
\begin{figure*}[ht]
\centering
    \includegraphics[width=1\linewidth]{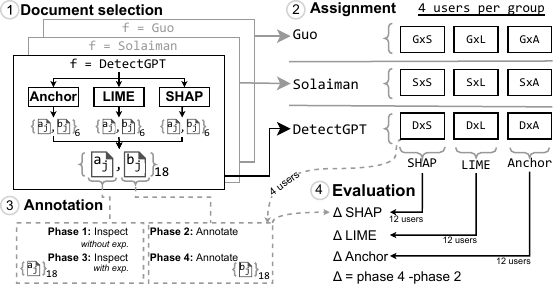}%
    \caption{Setup of the user study}
    \label{flowchartuserstudy}
\end{figure*}
\noindent Users can only apply observations about detector behavior in the annotation phases (2 and 4) if they have seen similar cases in the teaching phases (1 and 3).
Rather than using random documents, we construct two sets $\{a_j\}$ and $\{b_j\}$.
Each document $a_j$, to be shown in phases 1 and 3, has a corresponding document $b_j$ shown in phases 2 and 4. The explanations for the documents in a pair $(a_j,b_j)$ should be \textit{sufficiently similar} so that the task is feasible for human annotators.

In the first step, all possible combinations of documents in the dataset are sorted by their cosine similarity in a bag-of-words encoding of the most salient features (as in \citealp{ribeiro_why_2016}).
A second step aims to maintain sufficient diversity across pairs: 
Among the top-k most similar pairs, the $n$ pairs that maximize coverage are chosen, defined here as the number of features with a non-zero importance score in the global encoding.

For rule-based explanations, this strategy cannot be applied, as we lack an appropriate similarity metric for this type of rule.
Pairs are found by testing for set-equality: For a given document $a_j$, all other documents in the dataset that share an Anchor with $a_j$, and have $f(d_i) = f(a_j)$ are collected.
Anchors can be as short as one token long, and documents therefore might only overlap in one word with this strategy.  
A second step addresses this: If multiple documents share an Anchor with $a_j$, only the document with the highest Jaccard index with $a_j$ is kept. This is done so that the two documents share as similar of a context as possible. Finally, $n$ pairs are selected at random from these candidates. 

Note that these selection strategies will yield different sets for different pairings of detectors and explanation methods.
The experiments are conducted with three sets of pairs, one per detector (Figure \ref{flowchartuserstudy}).
Of the 18 pairs in a set, 6 will be determined by each explanation method. Document- and explanation similarity for the three sets are reported in tables \ref{results_document_selection_document_similarity} and \ref{results_document_selection_explanation_similarity}.
The datasets are balanced in respect to the detectors' predictions, to eliminate counting examples in phases 1 or 3 to infer the number of machine- and human-documents in phases 2 or 4 as a strategy \cite{hase_evaluating_2020}.
\paragraph{Assignment.}

In total, there are 9 different explanation method-detector combinations. 
One user only sees explanations from one explanation method for one detector. Users who are assigned the same detector, but different explanation methods, see the same set of documents. Those who are assigned the same explanation method, but different detectors do not. 
\paragraph{Evaluation.} We report the average change in user accuracy per explanation method. Whether showing explanations leads to a significant increase in performance is assessed with McNemar's test \cite{mcnemar_note_1947}.

Tables \ref{results_document_selection_document_similarity} and \ref{results_document_selection_explanation_similarity} report the document- and explanation similarity of pairs obtained with the proposed selection strategy. Pairs of explanations are significantly more similar than random selections, while not featuring overly similar documents. 

\FloatBarrier
\onecolumn
\begin{table*}[ht!]
\centering
\footnotesize

\begin{tabular}{l|l|rrr}
 & \textbf{Set}  & \textbf{Method}& \textbf{Random} & \textbf{Increase} \\
\hline
Jaccard & Solaiman & 0.17 & 0.12 & \bfseries 0.05 \\
 & Guo & 0.14 & 0.12 & \bfseries 0.02 \\
 & DetectGPT & 0.14 & 0.11 & \bfseries 0.03 \\
\hline
Cosine TF-IDF & Solaiman & 0.12 & 0.08 & \bfseries 0.05 \\
 & Guo & 0.12 & 0.09 & \bfseries 0.04 \\
 & DetectGPT & 0.12 & 0.08 & \bfseries 0.04 \\
\end{tabular}
\caption{Document Similarity between pairs $(a_j,b_j)$ in the datasets for the user study against the mean similarity of 10 random selections ($p < 0.05$ bold). Cosine similarity is given as the mean of SHAP and LIME}
\label{results_document_selection_document_similarity}
\end{table*}
\begin{table*}[ht!]
\centering
\footnotesize

\begin{tabular}{l|l|rrr}
 & \textbf{Set}  & \textbf{Method}& \textbf{Random} & \textbf{Gain} \\
\hline
Cosine Sim FI-Features & Solaiman & 0.31 & 0.17 & \bfseries 0.15 \\
 & Guo & 0.38 & 0.22 & \bfseries 0.16 \\
 & DetectGPT & 0.24 & 0.11 & \bfseries 0.14 \\
\hline
\# Matching Anchors  & Solaiman & 0.33 & 0.09 & \bfseries 0.24 \\
 & Guo & 0.89 & 0.42 & \bfseries 0.47 \\
 & DetectGPT & 0.28 & 0.06 & \bfseries 0.22 \\
\end{tabular}

\caption{Explanation Similarity between pairs $(a_j,b_j)$ in the datasets for the user study against the mean similarity of 10 random selections ($p < 0.05$ bold). Cosine similarity is given as the mean of SHAP and LIME}
\label{results_document_selection_explanation_similarity}
\end{table*}

\begin{figure}[ht!]
    \fbox{%
    \parbox{\linewidth}{%
    \small
    For each document, please also rate to what extent you agree with these statements:
    \begin{itemize}
    \small
        \item[Q1] From the explanation, I \textbf{understand why} the detector decided the way it did for this document.

        Select \textit{agree} or \textit{strongly agree} if you think the visualization presents sufficient evidence for why the detector decided the way it did in this specific case.

        Select \textit{disagree} or \textit{strongly disagree} if you can't figure out why the detector decided the way it did.
        \item[Q2] From the explanation, I now better \textbf{understand how} the detector works.

        Select \textit{agree} or \textit{strongly agree} if you think this explanation increased your understanding of how the detector reasons.

        Select \textit{disagree} or \textit{strongly disagree} if you don't.
        \item[Q3] The information from this explanation \textbf{will help me} predict the detector's behaviour.
        
        Select \textit{agree} or \textit{strongly agree} if you think you could apply this information to the documents you labelled in the previous phase. You will do so in the next phase.
        
        Select \textit{disagree} or \textit{strongly disagree} if you don't.
    \end{itemize}
    }%
    }
    \caption{Instructions for the Likert scale items shown in Phase 3 adapted from \citet{hoffman_metrics_2019}}
    \label{lickert-appendix}
\end{figure}

\begin{figure}[ht!]
\centering

    \includegraphics[width=0.8\textwidth]{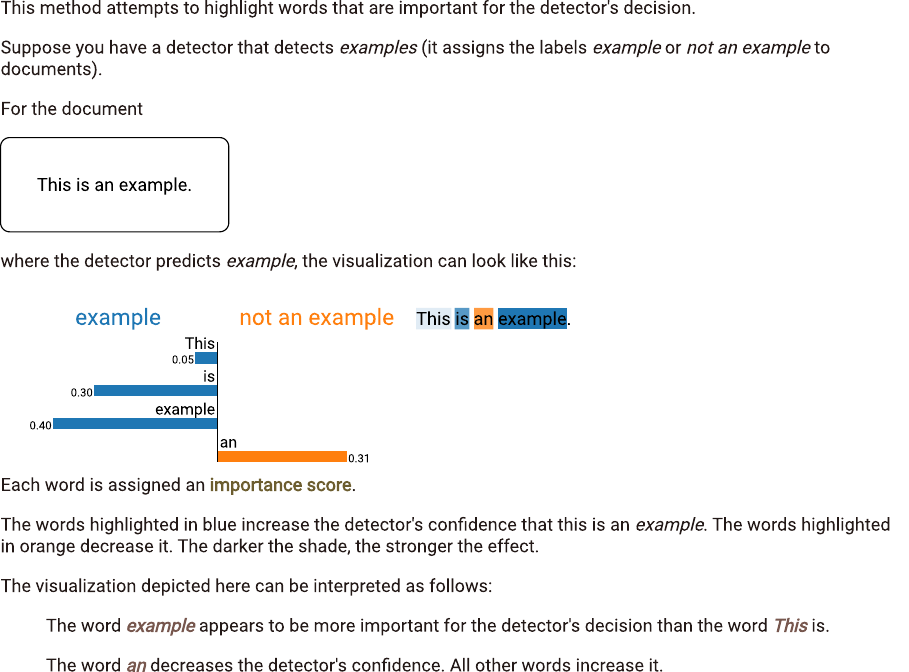}

\caption{Instructions on LIME shown in Phase 3}
\label{lime_exp2}
\end{figure}

\begin{figure}[ht!]
\centering

    \includegraphics[width=0.6\textwidth]{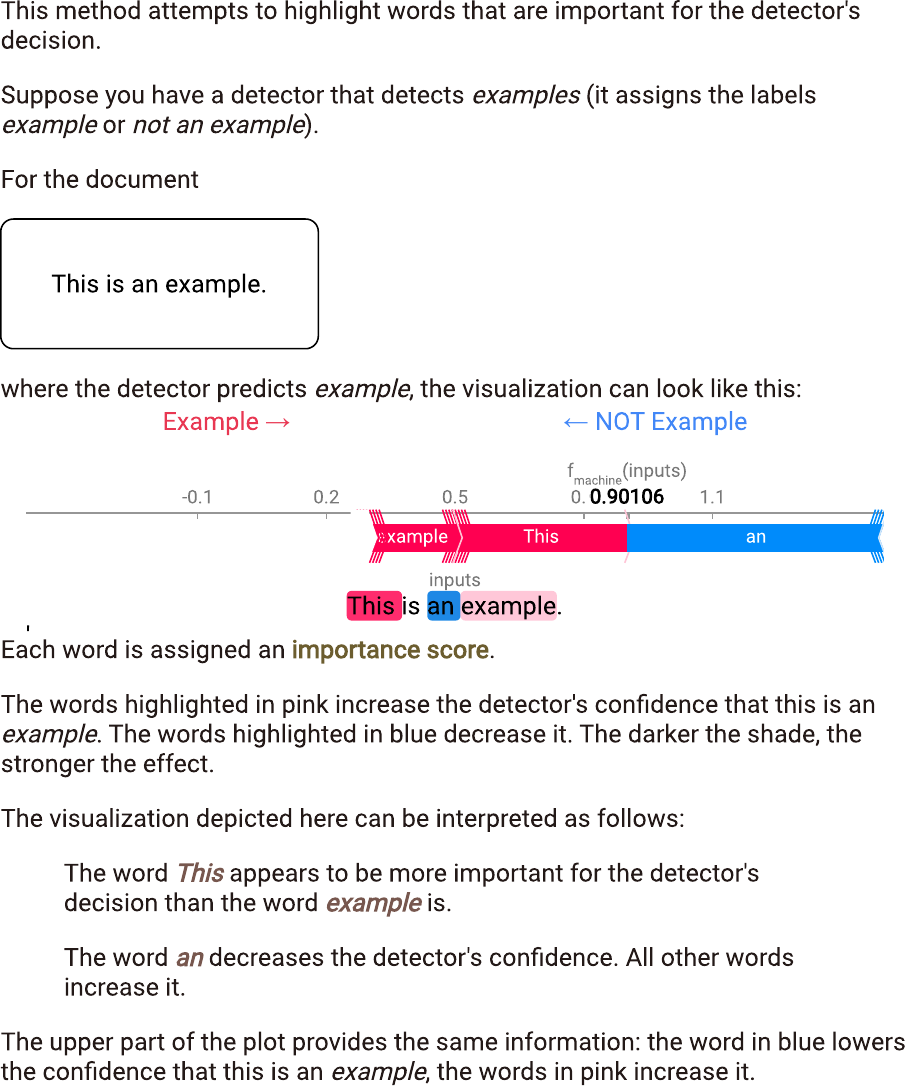}

\caption{Instructions on SHAP shown in Phase 3}
\label{shap_exp2}
\end{figure}

\begin{figure}[ht!]
\centering

    \includegraphics[width=0.8\textwidth]{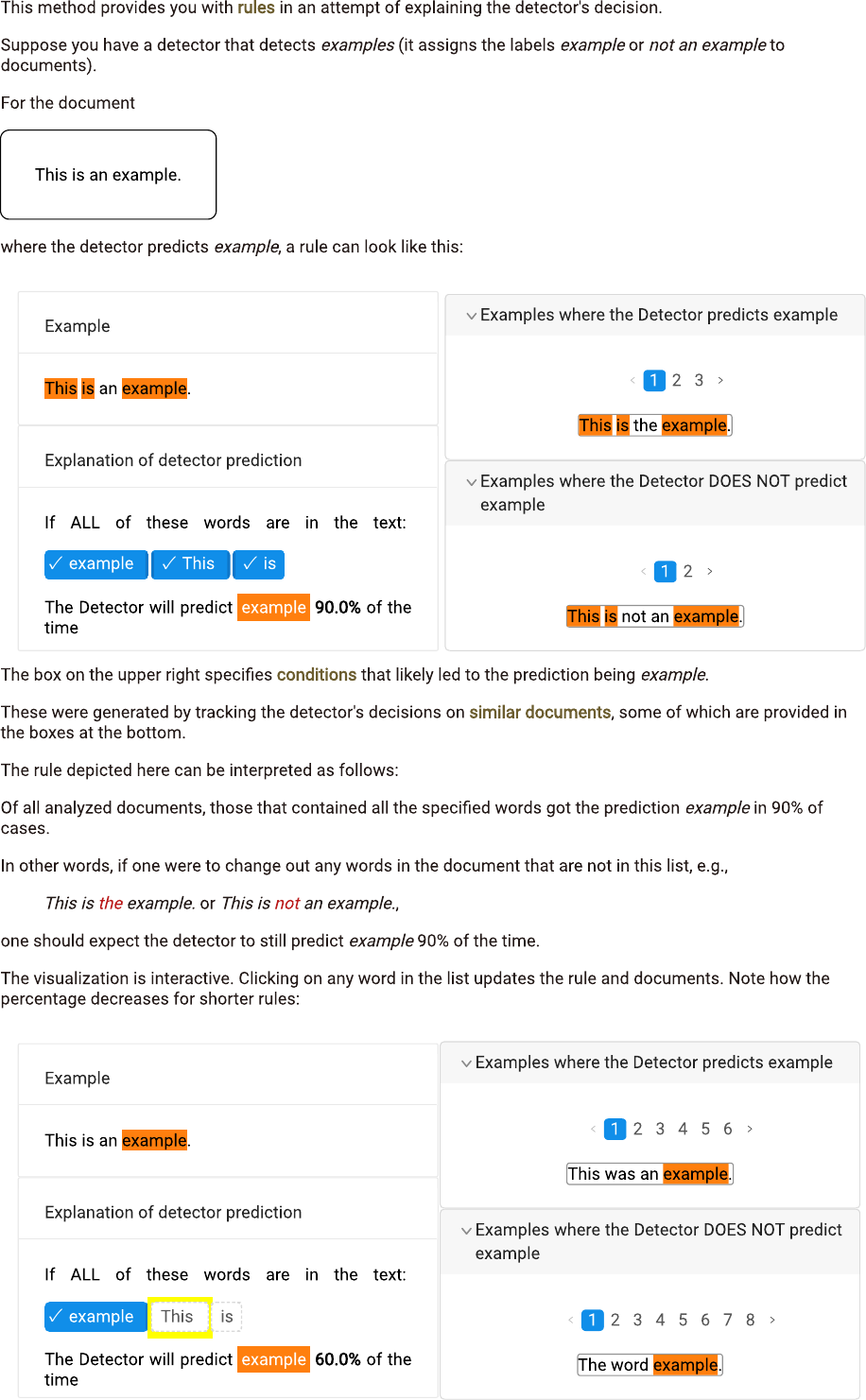}

\caption{Instructions on Anchor shown in Phase 3}
\label{anchor_exp2}
\end{figure}

\begin{figure}[ht!]
\centering
\includegraphics[width=\textwidth]{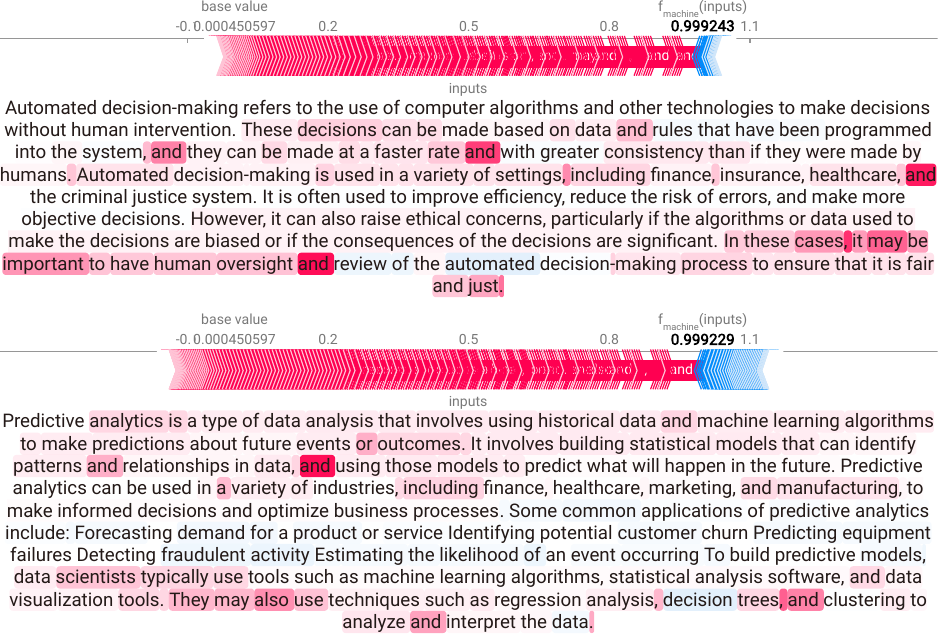}
\caption{Set Guo, selected by SHAP; $y_i=f(d_i)=\text{machine}$. Users only see explanations for the first document of each pair and annotate the second one}
\label{pair_shap}
\end{figure}
\vspace{1em}

\begin{figure}[ht!]
\centering
\includegraphics[width=1\linewidth]{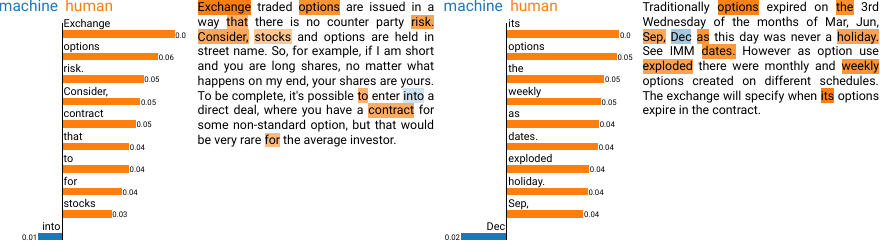}
\caption{Set DetectGPT, selected by LIME; $y_i=f(d_i)=\text{human}$. Users only see explanations for the first document of each pair and annotate the second one}
\label{pair_lime}
\end{figure}

\begin{figure}[ht!]
\centering
\includegraphics[width=\textwidth]{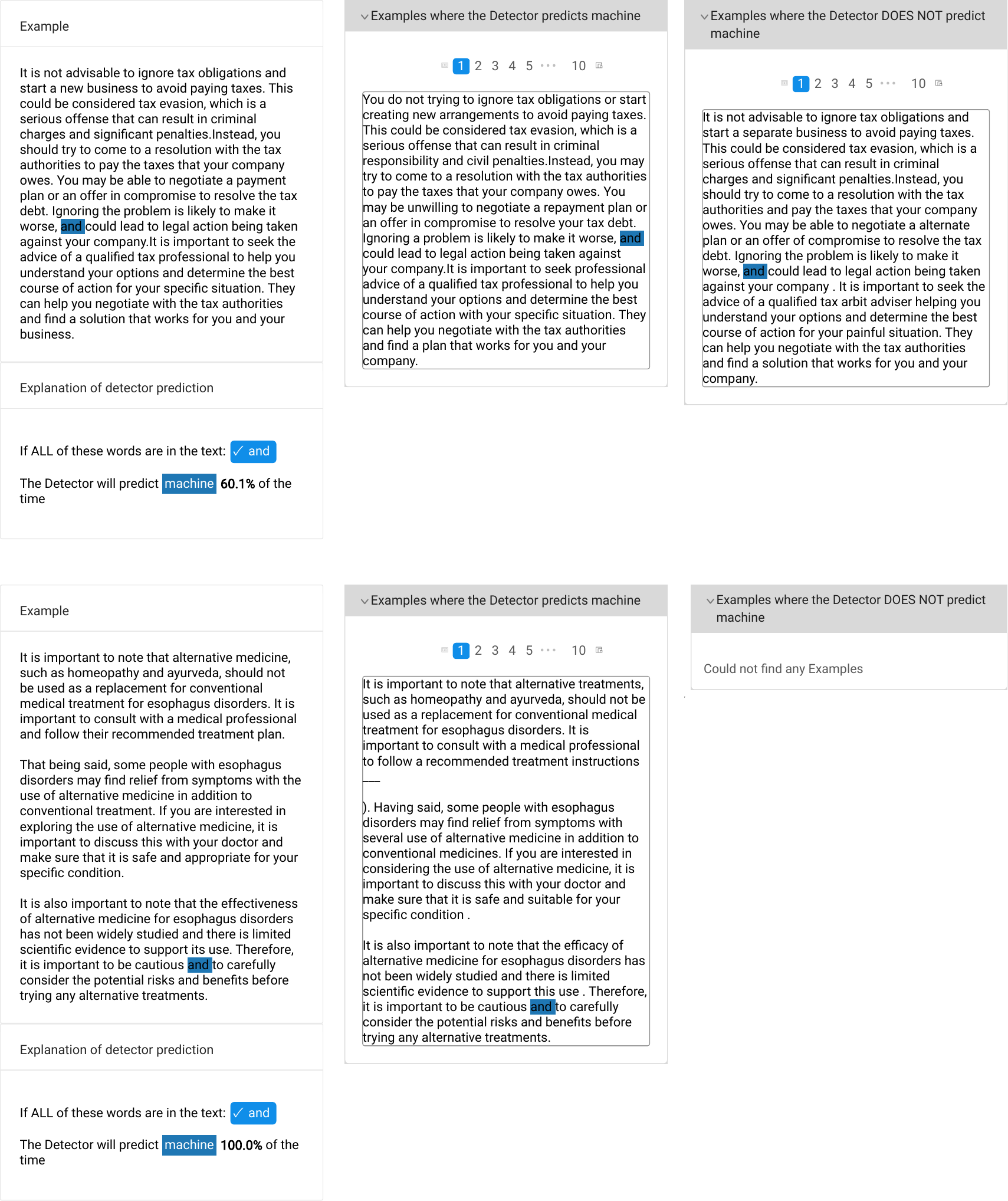}

\caption{Set Solaiman, selected by Anchor; $y_i=f(d_i)=\text{machine}$. Users only see explanations for the first document of each pair and annotate the second one}
\label{pair_anchor}
\end{figure}
\FloatBarrier
\section{Complementary results}\label{S2_Appendix}
\begin{figure}[ht!]
    \begin{center}
    
    \resizebox{0.8\textwidth}{!}{
        \input{masking.pgf}
    }
    \end{center}
    \caption{Perturbation Strategy: Shift in detector output at different percentages of tokens randomly removed or replaced. The original dataset is balanced, the counts at 0\% of tokens removed serve as a baseline. Mean of 10 runs}
    \label{masking-figure}
\end{figure}

\begin{table}[ht!]
\centering
\footnotesize
\begin{tabular}{lllrl}
\hline
\textbf{Detector} & \textbf{Explainer} &   \textbf{Score} & \textbf{p} \\
\hline
Guo & LIME\  & 0.605 & 0.168 \\
\hline
& Random\  & 0.635 & 0.000 \\
\hline 
& Anchor\  & 0.681 & 0.000 \\
\hline
& SHAP\  & 0.812 &  \\
\hline
\hline
Solaiman & LIME\ & 0.402 & 0.004 \\
\hline
& Random\ & 0.484 & 0.003 \\
\hline
& Anchor\  & 0.492 & 0.000 \\
\hline
& SHAP\  & 0.631 &  \\
\hline
\hline
DetectGPT & Random\ & 0.577 & 0.000 \\
\hline
& Anchor\ & 0.596 & 0.116 \\
\hline
& LIME\ & 0.631 & 0.473 \\
\hline
& SHAP\ & 0.635 &  \\
\hline
\hline
\end{tabular}

\caption{Pointing Game: Per detector results. P-values from row-wise binominal tests}
\label{results_pointing_game_detector}
\end{table}
\begin{table}[ht!]
\centering
\footnotesize
\begin{tabular}{llr}
\hline
Explainer & Detector & $\alpha$ \\
\hline
SHAP & Guo & \bfseries 0.896 \\
\hline
 & Solaiman & 0.812 \\
\hline
 & DetectGPT & 0.081 \\
\hline
LIME & Guo & 0.478 \\
\hline
 & Solaiman & 0.439 \\
\hline
 & DetectGPT & 0.265 \\
\hline
Anchor & Guo & 0.367 \\
\hline
 & Solaiman & 0.135 \\
\hline
 & DetectGPT & 0.129 \\
\hline
Random & Solaiman & -0.137 \\
\hline
 & Guo & -0.137 \\
\hline
 & DetectGPT & -0.144 \\
\hline
\hline
\end{tabular}
\caption{Continuity: Per detector results}
\label{table_combined_continuity}
\end{table}
\begin{table}[ht]
\footnotesize
\centering
\begin{tabular}{llrrrr}
\hline
 &  & User Acc without & User Acc with & Change & p \\
\hline
LIME & Solaiman & 0.681 & 0.569 & -16.33\% & 0.057 \\
LIME & DetectGPT & 0.792 & 0.681 & -14.04\% & 0.134 \\
LIME & Guo & 0.750 & 0.681 & -9.26\% & 0.359 \\
Anchor & DetectGPT & 0.806 & 0.750 & -6.90\% & 0.289 \\
SHAP & DetectGPT & 0.819 & 0.806 & -1.69\% & 1.000 \\
SHAP & Solaiman & 0.667 & 0.681 & 2.08\% & 1.000 \\
Anchor & Solaiman & 0.569 & 0.583 & 2.44\% & 1.000 \\
Anchor & Guo & 0.708 & 0.764 & 7.84\% & 0.344 \\
SHAP & Guo & 0.778 & 0.847 & 8.93\% & 0.227 \\
\hline

\end{tabular}
\caption{Forward Simulation: Per group results.}
\label{faithfulnessperdetector}
\end{table}
\begin{table}[ht]
\footnotesize
\centering
\begin{tabular}{lllr}

\hline
DetectGPT & Anchor & Q1 & 2.15 \\

 &  & Q2 & 2.04 \\

 &  & Q3 & 2.11 \\
\hline
 & LIME & Q1 & 3.62 \\

 &  & Q2 & 2.97 \\

 &  & Q3 & 3.08 \\
\hline
 & SHAP & Q1 & 2.54 \\

 &  & Q2 & 2.56 \\

 &  & Q3 & 2.57 \\
\hline
\hline
Solaiman & Anchor & Q1 & 3.17 \\

 &  & Q2 & 3.18 \\

 &  & Q3 & 3.15 \\
\hline
 & LIME & Q1 & 4.00 \\

 &  & Q2 & 3.81 \\

 &  & Q3 & 3.65 \\
\hline
 & SHAP & Q1 & 3.33 \\

 &  & Q2 & 3.39 \\

 &  & Q3 & 3.29 \\
\hline
\hline
Guo & Anchor & Q1 & 2.39 \\

 &  & Q2 & 2.22 \\

 &  & Q3 & 2.26 \\
\hline
 & LIME & Q1 & 3.17 \\

 &  & Q2 & 3.32 \\

 &  & Q3 & 3.19 \\
\hline
 & SHAP & Q1 & 3.29 \\

 &  & Q2 & 2.64 \\

 &  & Q3 & 2.65 \\
\hline
\end{tabular}

\caption{Perceived Usefulness: Per group results. (3 = neutral, 5 = strongly agree)}
\label{usefulnessperdetector}
\end{table}

\begin{table}[ht!]
\footnotesize
\centering
\begin{tabular}{llr}
\hline
\textbf{Explainer} & \textbf{Detector} &  $\alpha$\\
\hline
SHAP & Solaiman & \bfseries 1.000 \\
\hline
 & Guo & 1.000\\
 \hline
 & DetectGPT & 0.084\\
\hline
LIME & Solaiman & 0.204 \\
\hline
 & Guo & 0.179 \\
 \hline
 & DetectGPT & 0.023 \\
\hline
Anchor & Solaiman & 0.135 \\
\hline
 & Guo & 0.316 \\
\hline
 & DetectGPT & 0.097 \\
\hline
Random  & Solaiman & -0.167 \\
\hline
& Guo & -0.167 \\
\hline
& DetectGPT & -0.167 \\
\hline
\hline
\end{tabular}
\caption{Consistency: Per detector results}
\label{table_combined_consistency}
\end{table}

\begin{figure}
\centering
\footnotesize
\addtolength{\tabcolsep}{-0.3em} 
\begin{tabular}{llllrrr}
 &  & \textbf{$f(d_i)$ → $f(d_i^\Omega)$} & \textbf{n} &\textbf{$c_\textit{intra}$} & \textbf{$c_\textit{inter}$} \\
 \hline
\textbf{DetectGPT}&  Random & h → m & 800 & 0.48 & 0.49 \\
\hline
 &  & m → h & 1990 & 0.50 & 0.49 \\
\hline
& LIME & h → m & 80 & 0.39 & 0.74 \\
\hline
 &  & m → h & 199 & 0.62 & 0.28 \\
 \hline
& Anchor & h → m & 80 & 0.60 & 0.71 \\
\hline
 &  & m → h & 199 & 0.36 & 0.56 \\
\hline
 & SHAP & h → m & 80 & 0.59 & 0.59 \\
\hline
 &  & m → h & 199 & 0.59 & 0.63 \\
\hline
\textbf{Guo}  & Random & h → m & 90  & 0.46 & 0.34 \\
\hline
 &  & m → h & 1530 & 0.49 & 0.50 \\
\hline
& LIME & h → m & 9  & 0.78 & 0.44 \\
\hline
 &  & m → h & 153 & 0.52 & 0.73 \\
 \hline
& Anchor & h → m & 9 & 1.00 & 0.22 \\
\hline
 &  & m → h & 153 & 0.66 & 0.01 \\
\hline
 & SHAP & h → m & 9 & 1.00 & 1.00 \\
\hline
 &  & m → h & 153 & 0.93 & 0.71 \\
\hline
Solaiman  & Random & h → m & 1420 & 0.48 & 0.51 \\
\hline
 &  & m → h & 1480 & 0.48 & 0.51 \\
\hline
& LIME & h → m & 142 & 0.79 & 0.90 \\
\hline
 &  & m → h & 148 & 0.56 & 0.48 \\
 \hline
& Anchor & h → m & 142 & 0.65 & 0.23 \\
\hline
 &  & m → h & 148 & 0.35 & 0.05 \\
\hline
 & SHAP & h → m & 142 & 0.97 & 0.94 \\
\hline
 &  & m → h & 148 & 0.88 & 0.97 \\
\hline
\end{tabular}
\caption{Contrastivity: Per detector results and data on the synthetic datasets}
\label{continuity_subfigure}
\end{figure}%

\begin{figure}
\centering
\resizebox{0.65\linewidth}{!}{\input{figure_contrastivity_label_flip_histogram.pgf}
}
\caption{Contrastivity: Fraction of tokens edited until $f(d_i^\Omega) \neq f(d_i)$}
\label{contrastivity-label-flip-histogram}
\end{figure}
\clearpage
\section{Detector Benchmark Results}\label{S3_Appendix}

\begin{table}[ht!]
\centering
\resizebox{\textwidth}{!}{%
\begin{tabular}{lrrrrrrrr}
\hline
 & \textbf{Acc} & \textbf{F1} & \textbf{AUC} & \textbf{TN} & \textbf{FP} & \textbf{FN} & \textbf{TP} &\textbf{ms/evaluation} \\
\hline
DetectGPT GPT-2 @100 samples & 0.502 & 0.000 & 0.500 & 153 & 0 & 152 & 0 & 15808 \\
\hline
DetectGPT pythia-70m @100 samples & 0.705 & 0.579 & 0.704 & 153 & 0 & 90 & 62 & 6391 \\
\hline
DetectGPT pythia-70m (this paper)& 0.744 & 0.664 & 0.743 & 150 & 3 & 75 & 77 & 898 \\
\hline
Solaiman & 0.921 & 0.922 & 0.921 & 139 & 14 & 10 & 142 & 19 \\
\hline
Guo & 0.990 & 0.990 & 0.990 & 153 & 0 & 3 & 149 & 18 \\
\hline
\hline
\end{tabular}

}
\caption{{\bf Performance of the detectors.} Note that one could obtain better results for DetectGPT when using GPT-2 by adjusting the clas\-sification threshold.}
\label{table_performance}
\end{table}

\end{document}